\pdfoutput=1
\documentclass[11pt]{article}

\usepackage[preprint]{acl}
\usepackage{times}
\usepackage{latexsym}
\usepackage{enumitem}

\usepackage[T1]{fontenc}
\usepackage[utf8]{inputenc}

\usepackage{microtype}

\usepackage{inconsolata}

\usepackage{graphicx}
\usepackage{times}
\usepackage{listings}
\usepackage{multirow}
\usepackage{soul} % for strikethrough
\usepackage{subcaption}
\usepackage{graphics}
\usepackage{xspace}
\usepackage{amsmath}
\usepackage{amssymb}  % for checkmarks.
\usepackage{adjustbox}
\usepackage{makecell}
\usepackage{pifont}
\usepackage{tabularx}
\usepackage{xcolor}
\usepackage{booktabs}

\graphicspath{{./figs/}}

\newcommand{\iso}{{\textsc{ISO-Bench}}\xspace}
\newcommand{\catbench}{{\textsc{CaT-Bench}}\xspace}
\newcommand{\dep}{\textsc{Dep}\xspace}
\newcommand{\ndep}{\textsc{NonDep}\xspace}
\newcommand{\mavg}{\textsc{Macro Avg}\xspace}

\newcommand{\llms}{\textsc{LLM}s\xspace}
\newcommand{\vlm}{\textsc{VLM}\xspace}
\newcommand{\vlms}{\textsc{VLM}s\xspace}

\newcommand{\gfomni}{\texttt{4o}\xspace}
\newcommand{\gfmini}{\texttt{4o-mini}\xspace}
\newcommand{\sonnet}{\texttt{Sonnet}\xspace}
\newcommand{\othree}{\texttt{o3}\xspace}
\newcommand{\ofour}{\texttt{o4-mini}\xspace}

\newcommand{\llava}{\texttt{Llava-1.5}\xspace}
\newcommand{\llama}{\texttt{Llama-3.2}\xspace}
\newcommand{\pali}{\texttt{PaliGemma2}\xspace}

\newcommand{\qwen}{\texttt{Qwen2.5-VL}\xspace}
\newcommand{\ds}{\texttt{DeepSeek-VL2}\xspace}

\newcommand{\aexpt}{\textsc{A}\xspace}
\newcommand{\eaexpt}{\textsc{E}$\rightarrow$\textsc{A}\xspace}

\newcommand{\squishlist}{
  \begin{list}{$\bullet$}
    { \setlength{\itemsep}{0pt}      \setlength{\parsep}{3pt}
      \setlength{\topsep}{3pt}       \setlength{\partopsep}{0pt}
      \setlength{\leftmargin}{1.5em} \setlength{\labelwidth}{1em}
      \setlength{\labelsep}{0.5em} } }
\newcommand{\reallysquishlist}{
  \begin{list}{$\bullet$}
    { \setlength{\itemsep}{0pt}    \setlength{\parsep}{0pt}
      \setlength{\topsep}{0pt}     \setlength{\partopsep}{0pt}
      \setlength{\leftmargin}{0.2em} \setlength{\labelwidth}{0.2em}
      \setlength{\labelsep}{0.2em} } }

 \newcommand{\squishend}{
     \end{list} 
 }

\usepackage{todonotes}
\title{\iso: Benchmarking Multimodal Causal Reasoning in Visual–Language Models through Procedural Plans}

\author{Ananya Sadana \\
  Stony Brook University \\
  \texttt{ananya.sadana@stonybrook.edu} \\\And
  Yash Kumar Lal \\
  Stony Brook University \\
  ylal@stonybrook.edu \\\And
  Jiawei Zhou \\
  Stony Brook University \\
  \texttt{jiawei.zhou1@stonybrook.edu} \\}

\begin{document}
\maketitle
\begin{abstract}
Understanding causal relationships across modalities is a core challenge for multimodal models operating in real-world environments. We introduce \iso, a benchmark for evaluating whether models can infer causal dependencies between visual observations and procedural text. Each example presents an image of a task step and a text snippet from a plan, with the goal of deciding whether the visual step occurs before or after the referenced text step. 
Evaluation results on ten frontier vision-language models show underwhelming performance: the best zero-shot F1 is only 0.57, and chain-of-thought reasoning yields only modest gains (up to 0.62 F1), largely behind humans (0.98 F1).
Our analysis further highlights concrete directions for improving causal understanding in multimodal models.
\end{abstract}

\section{Introduction}

Complex visual reasoning is critical for advancing multimodal intelligence, particularly in tasks involving planning and execution in the physical world. In such contexts, visual information plays a central role in how humans interpret, follow, and carry out plans. For instance, assembling furniture with a manual or cooking from a recipe requires understanding the logical and temporal structure of the plan, grounding visual observations within this structure, and reasoning about current and next steps by integrating information across modalities.

\begin{figure}[!t]
    \centering
    \includegraphics[width=\columnwidth]{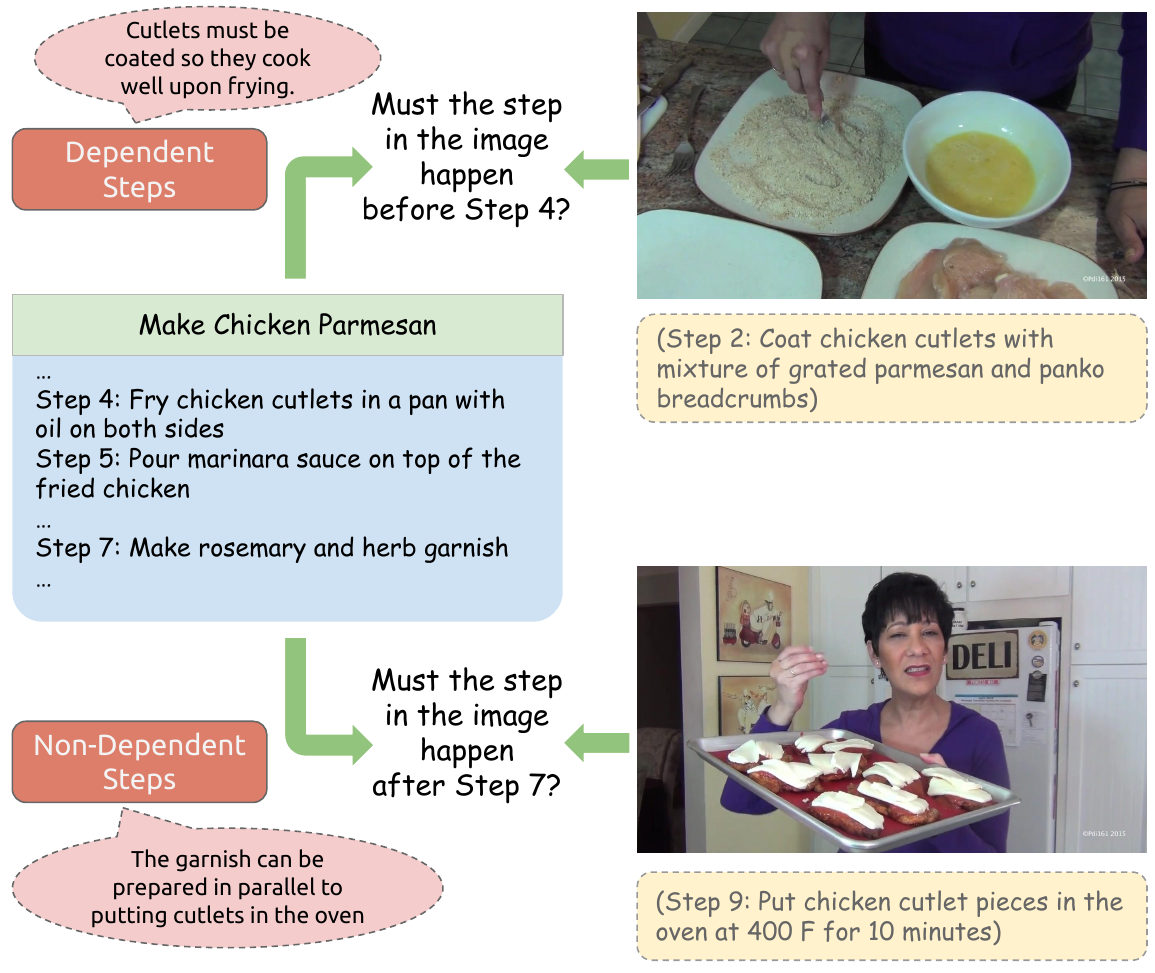}
    \caption{An example plan, visual states, and their causal reasoning. To determine whether the step in the top image must occur before Step 4, one must recognize it as coating chicken cutlets and infer that frying depends on it. In contrast, the bottom image shows cutlets being placed in the oven—a step independent of Step 7 (garnish preparation), which can occur in parallel.}
    \label{fig:iso_example}
\end{figure}

As vision-language models (VLMs) continue to improve, a key question emerges: can these models perceive and reason about multimodal plans that unfold in real-world visual contexts? In this work, we focus on a specific and challenging aspect of this problem—\textit{causal reasoning across modalities}. Given a plan in text and a visual observation (e.g., an image showing the status of a task in \autoref{fig:iso_example}), we investigate whether a model can understand the causal relationship between steps conveyed through different modalities.
This line of inquiry forms the foundation for more advanced reasoning capabilities, including video understanding, grounded world modeling, and embodied AI, where agents must interpret visual states to execute plans.

While existing benchmarks in visual intelligence often focus on image or video question answering \citep{yue2024mmmu, fu2025video, fang2024mmbench}, they typically lack the procedural and temporal structure needed for physical planning. In contrast, plan reasoning benchmarks are usually text-only \citep{valmeekam-etal-2023-planbench, lin-etal-2020-recipe}, focusing on state changes \citep{Bosselut2018SimulatingAD} or action sequences \citep{donatelli-etal-2021-aligning}. Yet real-world plans often rely on visual context and demand reasoning across modalities. Inferring dependencies in such settings requires not only perception and grounding, but also understanding causal relations like preconditions and effects between steps.

To address this gap, we introduce \iso (Image-State Ordering Benchmark), a new evaluation set to study whether models can reason about causal relationships across modalities in procedural plans. Built from two instructional video datasets—YouCook2 \cite{zhou-youcook2-2018} and CrossTask \cite{Zhukov2019}—we manually construct and annotate a diverse set of examples spanning domains like cooking, car maintenance, crafting, and woodworking. Each example includes a text snippet, an image depicting a step, and a question asking whether the image step occurs before or after the textual steps—directly probing multimodal causal understanding.

We evaluate $10$ state-of-the-art multimodal models, both open-source and proprietary, on \iso\footnote{\iso is available at \url{https://huggingface.co/datasets/StonyBrookNLP/ISO-Bench}}. Surprisingly, even the strongest models perform near chance. Reasoning-based chain-of-thought \citep{wei-etal-2022-chain} yields only marginal gains, despite its effectiveness in other reasoning tasks \citep{camburu-etal-2018-esnli, kumar-talukdar-2020-nile}. These results reveal a notable gap in current \vlm capabilities and position multimodal causal reasoning as a concrete challenge for future research.

\section{Related Work}
\label{sec:related_work}

Prior work on plan understanding has examined various aspects such as entity state tracking \cite{Bosselut2018SimulatingAD, henaff2017tracking}, action prediction \cite{lin-etal-2020-recipe, donatelli-etal-2021-aligning}, and next-step inference \cite{zellers-etal-2019-hellaswag, zhang-etal-2020-analogous}. XPAD \cite{dalvi-etal-2019-everything} focuses on scientific processes with action-dependency prediction, while PizzaCommonsense \cite{diallo-etal-2024-pizzacommonsense} enriches cooking plans with fine-grained intermediate steps. PlanBench \cite{valmeekam-planbench-2023} and NaturalPlan \cite{zheng-naturalplan-2024} explore classical and constraint-based planning, showing that LLMs often struggle to produce valid action sequences.

Several benchmarks evaluate models' ability to reason about procedural dependencies, but primarily in the text modality. For example, \catbench \cite{lal-catbench-2024} and \citet{kiddon-etal-2015-mise} focus on dependency prediction in cooking recipes, while CREPE \cite{zhang-etal-2023-causal} probes comparative causal judgments in procedures.

On the multimodal side, RecipeQA \cite{yagcioglu-recipeqa-2018} aligns text with illustrative images, and TOMATO \cite{shangguan-tomato-2024} evaluates temporal reasoning from videos. COMKitchens \cite{comkitchens_eccv2024} and MM-ReS \cite{pan-etal-multi-2020} provide datasets with structured workflow annotations. However, these benchmarks do not directly assess causal reasoning across modalities.
\iso fills this gap by testing whether models can integrate visual and textual information to infer temporal dependencies within instructional plans.\footnote{Please find more detailed related work in Appendix~\ref{appendix:related-work}.}

\begin{table*}[!t]
\centering
\begin{tabular}{ c  c  r  r  r  r  r  r  r  r  r }
\toprule
\multicolumn{2}{l}{\multirow{2}{*}{}} & \multicolumn{3}{c}{\dep} & \multicolumn{3}{c}{\ndep} & \multicolumn{3}{c}{\mavg} \\
\midrule
\multicolumn{1}{c}{} &  & \multicolumn{1}{c}{P} & \multicolumn{1}{c}{R} & \multicolumn{1}{c}{F} & \multicolumn{1}{c}{P} & \multicolumn{1}{c}{R} & \multicolumn{1}{c}{F} & \multicolumn{1}{c}{P} & \multicolumn{1}{c}{R} & \multicolumn{1}{c}{F} \\
\midrule
\multirow{2}{*}{\ds} & \eaexpt & 0.57 & 0.07 & 0.12 & 0.50 & 0.95 & 0.66 & 0.54 & 0.51 & 0.39 \\
 & \aexpt & 0.52 & 0.04 & 0.07 & 0.50 & 0.97 & 0.66 & 0.51 & 0.50 & 0.36 \\
\midrule
\multirow{2}{*}{\llava} & \eaexpt & 0.51 & 0.56 & 0.53 & 0.51 & 0.45 & 0.47 & 0.51 & 0.50 & 0.50 \\
 & \aexpt & 0.50 & 0.96 & 0.66 & 0.54 & 0.05 & 0.09 & 0.52 & 0.51 & 0.52 \\
\midrule
\multirow{2}{*}{\qwen} & \eaexpt & 0.53 & 0.37 & 0.43 & 0.56 & 0.26 & 0.35 & 0.54 & 0.31 & 0.39 \\
 & \aexpt & 0.56 & 0.50 & 0.53 & 0.54 & 0.59 & 0.57 & 0.55 & 0.55 & 0.55 \\
\midrule
\multirow{2}{*}{\pali} & \eaexpt & 0.51 & 0.24 & 0.33 & 0.50 & 0.77 & 0.61 & 0.50 & 0.50 & 0.47 \\
 & \aexpt & 0.57 & 0.17 & 0.26 & 0.57 & 0.87 & 0.64 & 0.54 & 0.52 & 0.53 \\
\midrule
\multirow{2}{*}{\llama} & \eaexpt & 0.51 & 0.71 & 0.59 & 0.58 & 0.13 & 0.22 & 0.54 & 0.42 & 0.41 \\
 & \aexpt & 0.44 & 0.01 & 0.02 & 0.50 & 0.99 & 0.66 & 0.47 & 0.50 & 0.34 \\
\midrule
\multirow{2}{*}{\gfmini} & \eaexpt & 0.54 & 0.79 & 0.64 & 0.61 & 0.33 & 0.42 & 0.57 & 0.56 & 0.53 \\
 & \aexpt & 0.53 & 0.85 & 0.65 & 0.62 & 0.24 & 0.34 & 0.57 & 0.55 & 0.50 \\
\midrule
\multirow{2}{*}{\gfomni} & \eaexpt & 0.58 & 0.67 & 0.63 & 0.61 & 0.52 & 0.56 & 0.60 & 0.60 & 0.59 \\
 & \aexpt & 0.56 & 0.72 & 0.63 & 0.61 & 0.43 & 0.50 & 0.58 & 0.58 & 0.57 \\
\midrule
\multirow{2}{*}{\sonnet} & \eaexpt & 0.54 & 0.76 & 0.63 & 0.60 & 0.35 & 0.44 & 0.57 & 0.56 & 0.54 \\
 & \aexpt & 0.55 & 0.65 & 0.60 & 0.58 & 0.27 & 0.37 & 0.57 & 0.46 & 0.48 \\
\midrule
\othree & \eaexpt & 0.61 & 0.61 & 0.62 & 0.62 & 0.61 & 0.61 & 0.62 & 0.62 & 0.62 \\
\midrule
\ofour & \eaexpt & 0.58 & 0.73 & 0.65 & 0.63 & 0.46 & 0.54 & 0.61 & 0.60 & 0.59 \\
\midrule
Human* & - & 0.97 & 0.98 & 0.98 & 0.98 & 0.97 & 0.97 & 0.98 & 0.98 & 0.98 \\
\bottomrule
\end{tabular}
\caption{Performance of all models on  when just providing an answer \aexpt and when also explaining that answer \eaexpt.
We report per-label as well as weighted macro average precision, recall and F1 score.
We report human baseline numbers calculated on a subset of 200 random instances of \iso.}
\label{tab:main_results}
\end{table*}

\section{Benchmark Construction}
\label{sec:data}

Understanding and following instructional plans requires the ability to comprehend both textual steps and corresponding visual states, as well as the dependencies between them. We focus on \emph{temporal and causal dependencies} across modalities: if the effects of one step satisfy the preconditions of another, the former must occur before the latter. Recognizing such dependencies involves reasoning over actions shown in images and inferring preconditions, causes, sub-goals, and effects.

To construct \iso, we repurpose two instructional video datasets, YouCook2 \citep{zhou-youcook2-2018} and CrossTask \citep{Zhukov2019}, covering domains like cooking, woodworking, and more. From each annotated step in the videos, we extract a single frame at the midpoint. For each video and plan, we create text snippets by selecting the first or last $k \in \{2,3,4\}$ steps of a plan and pair it with an image from another step in the same plan.
When the image comes from a later step, we frame the question as: \emph{Must the step in the image happen after the snippet?} Conversely, if the image is from an earlier step, we ask whether it must happen \emph{before}.

Each (snippet, image) pair is annotated with a binary yes/no label indicating whether there is a causal dependence between the image step and any step in the text. This yields two types of instances: \dep, where such a dependence exists (ground truth is ``yes''), and \ndep, where no causal relationship is present (ground truth is ``no''). 

\iso contains 764 examples, roughly evenly split between the two types. The task challenges models to decide whether the visual step occurs before or after the textual steps, probing their ability to reason about cross-modal procedural dependencies. Evaluation is based on per-class precision, recall, and F1 score. Dataset statistics appear in \autoref{tab:data} (Appendix~\ref{appsec:data_stats}), and data processing details are in Appendices~\ref{appendix:youcook2} and~\ref{appendix:crosstask}.

\section{Benchmarking Models on \iso}

We benchmark the performance of a variety of state-of-the-art models on \iso.

\subsection{Models and Setup}

We evaluate $10$ different vision-language models (\vlms).
Open-source models include llava-1.5-7b-hf (\llava), Llama-3.2-11B Vision Instruct (\llama), Qwen2.5-VL-32B (\qwen), paligemma2-10b-mix-448 (\pali) and deepseek-VL2-small (\ds), and proprietary models include gpt-4o-mini-2024-07-18 (\gfmini), gpt-4o-2024-11-20 (\gfomni), claude-3-5-sonnet-20241022 (\sonnet), OpenAI o3-2025-04-16 (\othree) and o4-mini-2025-04-16 (\ofour), covering a range of model sizes and capabilities.
As described in \autoref{appsec:human_baseline}, we also establish how well humans can perform this task.

We test models under two zero-shot settings: (i) direct answer generation (\aexpt), and (ii) explanation-augmented reasoning (\eaexpt), where models are prompted to generate an explanation before answering. Full model details, prompts, and evaluation setup are provided in Appendices~\ref{appsec:benchmark_models} and~\ref{appsec:prompts}.

\subsection{Results}

\autoref{tab:main_results} presents the performance of all the models in different settings for \iso. 
We present per-class (\dep and \ndep) 
precision, recall and F1 score as well as macro average metrics.
Humans are easily able to perform this task, and achieve very high F1 score on \iso.
We make two main observations on model performance.

\paragraph{Most \vlms perform poorly on \iso.}
We observe that most models perform poorly on the task (\aexpt rows in \autoref{tab:main_results}), with overall F1 scores only ranging from $\sim$0.3-0.5. 
Only one model (out of eight) stands out: \gfomni (0.57 F1). 
On the contrary, humans perform near perfectly, achieving 0.98 F1.

Models are clearly better at judging when there is a dependence between the step in the image and the steps in the text snippet.
Interestingly, we observe that \emph{most} models have high precision but low recall for \ndep.
Models can reliably identify only some of the \ndep dependencies.
Contrastingly, the best performing models attain high recall and low precision on \dep, indicating a bias towards assessing most data points as having a dependence.

\paragraph{Generating explanations only helps marginally.} 
The \eaexpt rows in \autoref{tab:main_results} represent results where models, given the image and text snippet, first perform step-by-step reasoning and then generate a decision on whether there is a dependence between the snippet and the image.
Results show that reasoning leads to improvements ($\sim$0.02-0.06) for closed-source models, and for \ds and \llama, but gains are rather small.
\ofour, trained to analyze and do reasoning over images, achieves the highest performance (0.62 F1 respectively).
For other open-source models, there is a notable drop in performance ($\sim$0.02-0.16) indicating a weakness in their CoT abilities.
We hypothesize that this is due to the limitations of instruction tuning in vision-language modeling where models are mainly finetuned to describe or analyze images, not produce reasoning chains across modalities. 

\section{Analysis}

We analyze the performance of the best models by different attributes of the questions and prompts.

\subsection{Reasoning as a function of Step Distance}

\begin{figure}[tb]
    \centering
    \includegraphics[width=\columnwidth]{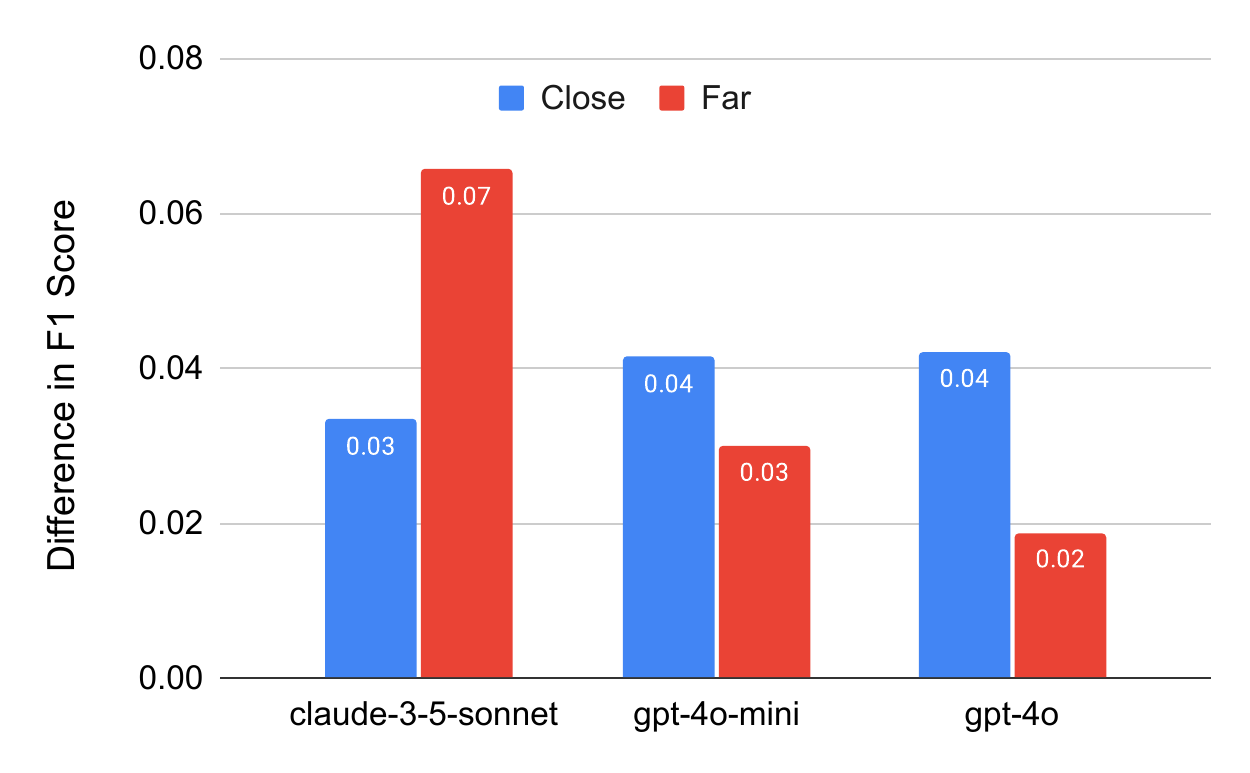}
    \caption{Difference in performance of models between (\eaexpt) and (\aexpt) settings split by the distance between the steps being asked about in the question.}
    \label{fig:dist_perf}
\end{figure}

We study how the distance between the step in the image and the closest step in the snippet impacts model (here, \gfmini, \gfomni, \sonnet) performance.
A question is said to be about \emph{close} steps if the distance between the step in the image ($step_i$) and the nearest step in the text snippet ($step_j$) is less than 3 ($|j - i| < 3$), and \emph{distant} otherwise.
\autoref{fig:dist_perf} shows the difference in F1 score between explaining then answering (\eaexpt) and just answering (\aexpt) with different models as a function of step distance.
As step distance increases, it should become harder to reason about whether there is a connection between them.
Explanations should provide larger gains for larger distance over smaller distance.
Surprisingly, this does not hold true for GPT models.

\subsection{Understanding Directional Dependencies}

\begin{figure}[tb]
    \centering
    \includegraphics[width=\columnwidth]{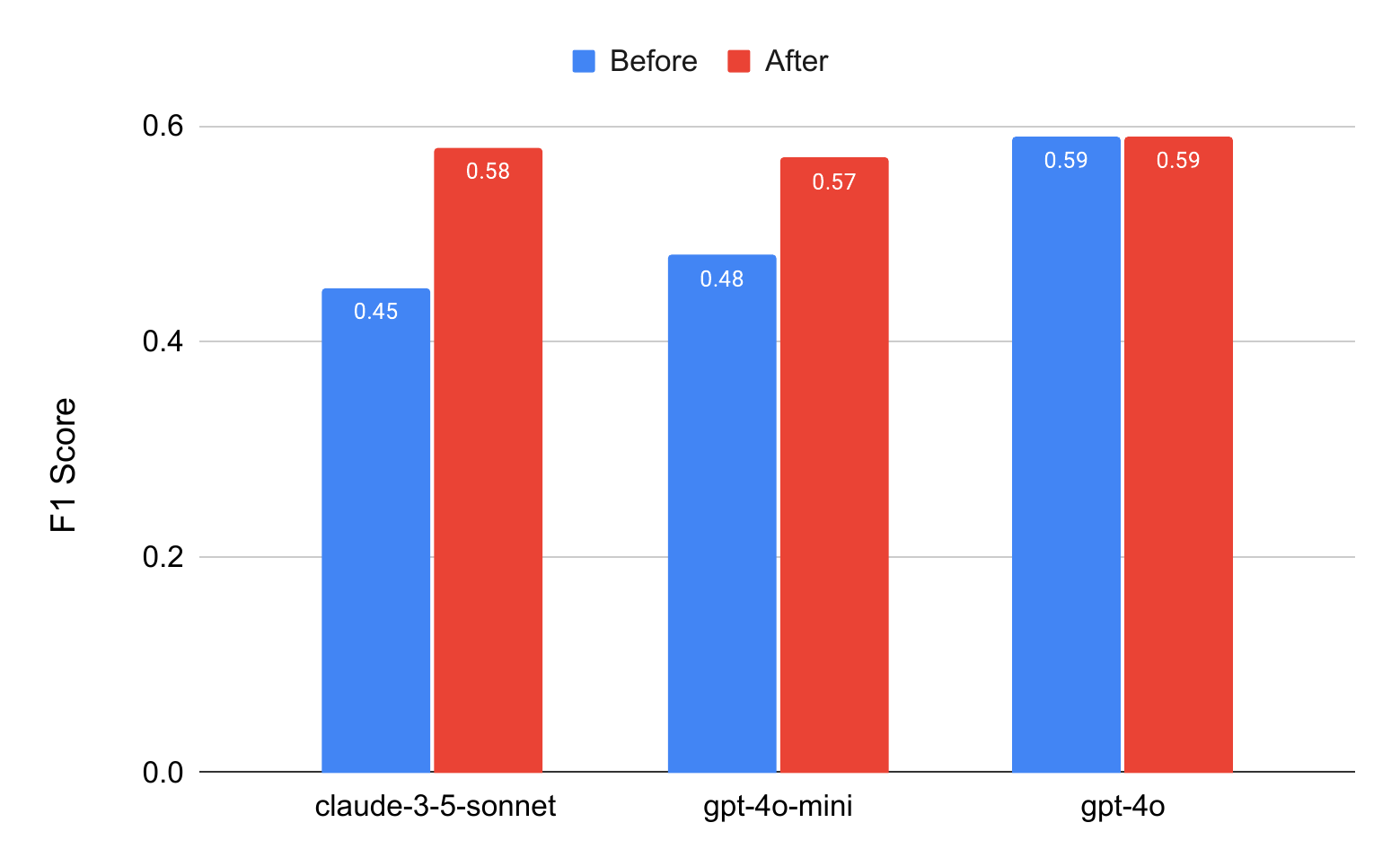}
    \caption{Performance of models (\aexpt) split by temporal relation type (\emph{before} and \emph{after}) in the question.}
    \label{fig:before_after}
\end{figure}

Next, we study how models handle questions about different aspects of the same pair of steps.
Typically, reasoning about whether a step must happen \emph{before} another requires reasoning about preconditions and causes, while understanding whether a step must happen \emph{after} another requires understanding the effects of any performed actions.
\autoref{fig:before_after} shows model performance (\aexpt setting) for these questions.
Most models are slightly better at reasoning about the effects of steps.
We hypothesize that this is because effects in plans may be more immediate and hence, would be easier to understand.

\subsection{Error Analysis}

To better understand model failures, we sample and analyze 100 explanations generated by \gfomni and \othree (\eaexpt) where the model produces an incorrect answer.
We identify 4 major types of errors:
\begin{itemize}
    \item Causal Reasoning (62\%) -- Models fail to understand that the step in the image describes either a precondition or an effect of (one of) the steps in the text snippet.
    \item Grounding (16\%) -- Often, models misidentify the step described in the image with another one not asked about in the question.
    \item Action Progression (14\%) -- Models misunderstand an action in progress in the image and assume that the action in question has already been completed when producing their answer.
    \item Perception (8\%) -- Models incorrectly identify items in the image leading to wrong reasoning.
\end{itemize}

\section{Conclusion}

We introduce \iso, a new benchmark to test whether multimodal models can reason about causal dependencies across visual and textual steps in procedural plans. Despite strong general capabilities, current \vlms perform poorly here, with only marginal gains from explanation-based prompting. Our work highlights a specific gap in multimodal causal reasoning and point to clear opportunities to advance model understanding of real-world plans.

\section*{Limitations}

Our work only investigates English-language documents (plans) and this limits the generalizability of our findings to other languages.

We benchmark a reasonably diverse set of \vlms. Currently, we cover several model families and models of varying sizes. However, due to the current fast-paced landscape of \vlm development, we are unable to cover all available \vlms. We will continue to evaluate more \vlms on \iso.

Due to difficulty in creating data, we were unable to create a ``dev'' set which could serve as a source of examples for few-shot experiments. 
Similarly, we are unable to perform fine-tuning or adapter based experiments on \iso.
We leave this exploration to future work.

\bibliography{custom,anthology}

\begin{thebibliography}{34}
\providecommand{\natexlab}[1]{#1}

\bibitem[{Bosselut et~al.(2018)Bosselut, Levy, Holtzman, Ennis, Fox, and Choi}]{Bosselut2018SimulatingAD}
Antoine Bosselut, Omer Levy, Ari Holtzman, Corin Ennis, Dieter Fox, and Yejin Choi. 2018.
\newblock Simulating action dynamics with neural process networks.
\newblock \emph{ICLR}.

\bibitem[{Camburu et~al.(2018)Camburu, Rockt\"{a}schel, Lukasiewicz, and Blunsom}]{camburu-etal-2018-esnli}
Oana-Maria Camburu, Tim Rockt\"{a}schel, Thomas Lukasiewicz, and Phil Blunsom. 2018.
\newblock \href {https://proceedings.neurips.cc/paper_files/paper/2018/file/4c7a167bb329bd92580a99ce422d6fa6-Paper.pdf} {e-snli: Natural language inference with natural language explanations}.
\newblock In \emph{Advances in Neural Information Processing Systems}, volume~31. Curran Associates, Inc.

\bibitem[{Dalvi et~al.(2018)Dalvi, Huang, Tandon, Yih, and Clark}]{dalvi-etal-2018-tracking}
Bhavana Dalvi, Lifu Huang, Niket Tandon, Wen-tau Yih, and Peter Clark. 2018.
\newblock \href {https://doi.org/10.18653/v1/N18-1144} {Tracking state changes in procedural text: a challenge dataset and models for process paragraph comprehension}.
\newblock In \emph{Proceedings of the 2018 Conference of the North {A}merican Chapter of the Association for Computational Linguistics: Human Language Technologies, Volume 1 (Long Papers)}, pages 1595--1604, New Orleans, Louisiana. Association for Computational Linguistics.

\bibitem[{Dalvi et~al.(2019)Dalvi, Tandon, Bosselut, Yih, and Clark}]{dalvi-etal-2019-everything}
Bhavana Dalvi, Niket Tandon, Antoine Bosselut, Wen-tau Yih, and Peter Clark. 2019.
\newblock \href {https://doi.org/10.18653/v1/D19-1457} {Everything happens for a reason: Discovering the purpose of actions in procedural text}.
\newblock In \emph{Proceedings of the 2019 Conference on Empirical Methods in Natural Language Processing and the 9th International Joint Conference on Natural Language Processing (EMNLP-IJCNLP)}, pages 4496--4505, Hong Kong, China. Association for Computational Linguistics.

\bibitem[{Diallo et~al.(2024)Diallo, Bikakis, Dickens, Hunter, and Miller}]{diallo-etal-2024-pizzacommonsense}
Aissatou Diallo, Antonis Bikakis, Luke Dickens, Anthony Hunter, and Rob Miller. 2024.
\newblock \href {https://doi.org/10.18653/v1/2024.findings-emnlp.728} {{P}izza{C}ommon{S}ense: A dataset for commonsense reasoning about intermediate steps in cooking recipes}.
\newblock In \emph{Findings of the Association for Computational Linguistics: EMNLP 2024}, pages 12482--12496, Miami, Florida, USA. Association for Computational Linguistics.

\bibitem[{Donatelli et~al.(2021)Donatelli, Schmidt, Biswas, K{\"o}hn, Zhai, and Koller}]{donatelli-etal-2021-aligning}
Lucia Donatelli, Theresa Schmidt, Debanjali Biswas, Arne K{\"o}hn, Fangzhou Zhai, and Alexander Koller. 2021.
\newblock \href {https://doi.org/10.18653/v1/2021.emnlp-main.554} {Aligning actions across recipe graphs}.
\newblock In \emph{Proceedings of the 2021 Conference on Empirical Methods in Natural Language Processing}, pages 6930--6942, Online and Punta Cana, Dominican Republic. Association for Computational Linguistics.

\bibitem[{Fang et~al.(2024)Fang, Mao, Duan, Zhao, Li, Lin, and Chen}]{fang2024mmbench}
Xinyu Fang, Kangrui Mao, Haodong Duan, Xiangyu Zhao, Yining Li, Dahua Lin, and Kai Chen. 2024.
\newblock Mmbench-video: A long-form multi-shot benchmark for holistic video understanding.
\newblock \emph{Advances in Neural Information Processing Systems}, 37:89098--89124.

\bibitem[{Fu et~al.(2025)Fu, Dai, Luo, Li, Ren, Zhang, Wang, Zhou, Shen, Zhang et~al.}]{fu2025video}
Chaoyou Fu, Yuhan Dai, Yongdong Luo, Lei Li, Shuhuai Ren, Renrui Zhang, Zihan Wang, Chenyu Zhou, Yunhang Shen, Mengdan Zhang, et~al. 2025.
\newblock Video-mme: The first-ever comprehensive evaluation benchmark of multi-modal llms in video analysis.
\newblock In \emph{Proceedings of the Computer Vision and Pattern Recognition Conference}, pages 24108--24118.

\bibitem[{Henaff et~al.(2017)Henaff, Weston, Szlam, Bordes, and LeCun}]{henaff2017tracking}
Mikael Henaff, Jason Weston, Arthur Szlam, Antoine Bordes, and Yann LeCun. 2017.
\newblock Tracking the world state with recurrent entity networks.
\newblock \emph{ICLR}.

\bibitem[{Kiddon et~al.(2015)Kiddon, Ponnuraj, Zettlemoyer, and Choi}]{kiddon-etal-2015-mise}
Chlo{\'e} Kiddon, Ganesa~Thandavam Ponnuraj, Luke Zettlemoyer, and Yejin Choi. 2015.
\newblock \href {https://doi.org/10.18653/v1/D15-1114} {Mise en place: Unsupervised interpretation of instructional recipes}.
\newblock In \emph{Proceedings of the 2015 Conference on Empirical Methods in Natural Language Processing}, pages 982--992, Lisbon, Portugal. Association for Computational Linguistics.

\bibitem[{Kumar and Talukdar(2020)}]{kumar-talukdar-2020-nile}
Sawan Kumar and Partha Talukdar. 2020.
\newblock \href {https://doi.org/10.18653/v1/2020.acl-main.771} {{NILE} : Natural language inference with faithful natural language explanations}.
\newblock In \emph{Proceedings of the 58th Annual Meeting of the Association for Computational Linguistics}, pages 8730--8742, Online. Association for Computational Linguistics.

\bibitem[{Lal et~al.(2024)Lal, Morita, Li, Rossi, and Lane}]{lal-catbench-2024}
Saksham Lal, Mizuki Morita, Yingya Li, Roberto Rossi, and Peter Lane. 2024.
\newblock \href {http://arxiv.org/abs/2406.15823} {{CaT{-}Bench}: Benchmarking language-model understanding of causal and temporal dependencies in plans}.
\newblock \emph{arXiv preprint}, arXiv:2406.15823.
\newblock To appear in EMNLP 2024.

\bibitem[{Lin et~al.(2020)Lin, Rao, Celikyilmaz, Nouri, Brockett, Dey, and Dolan}]{lin-etal-2020-recipe}
Angela Lin, Sudha Rao, Asli Celikyilmaz, Elnaz Nouri, Chris Brockett, Debadeepta Dey, and Bill Dolan. 2020.
\newblock \href {https://doi.org/10.18653/v1/2020.acl-main.440} {A recipe for creating multimodal aligned datasets for sequential tasks}.
\newblock In \emph{Proceedings of the 58th Annual Meeting of the Association for Computational Linguistics}, pages 4871--4884, Online. Association for Computational Linguistics.

\bibitem[{Liu et~al.(2023)Liu, Li, Wu, Li, Baldridge, and Seung}]{liu-llava-2023}
Haotian Liu, Chunyuan Li, Qingyang Wu, Yin Li, Jason Baldridge, and H.~Sebastian Seung. 2023.
\newblock \href {https://arxiv.org/abs/2310.03744} {{LLaVA{-}1.5}: Visual instruction tuning for large language-and-vision assistants}.
\newblock In \emph{Advances in Neural Information Processing Systems ({NeurIPS})}.

\bibitem[{Maeda et~al.(2024)Maeda, Hirasawa, Hashimoto, Harashima, Rybicki, Fukasawa, and Ushiku}]{comkitchens_eccv2024}
Koki Maeda, Tosho Hirasawa, Atsushi Hashimoto, Jun Harashima, Leszek Rybicki, Yusuke Fukasawa, and Yoshitaka Ushiku. 2024.
\newblock Com kitchens: An unedited overhead-view video dataset as a vision-language benchmark.
\newblock In \emph{Proceedings of the European Conference on Computer Vision}.

\bibitem[{Nguyen et~al.(2017)Nguyen, Nguyen, Chu, Thater, and Pinkal}]{nguyen-etal-2017-sequence}
Dai~Quoc Nguyen, Dat~Quoc Nguyen, Cuong~Xuan Chu, Stefan Thater, and Manfred Pinkal. 2017.
\newblock \href {https://aclanthology.org/I17-2007/} {Sequence to sequence learning for event prediction}.
\newblock In \emph{Proceedings of the Eighth International Joint Conference on Natural Language Processing (Volume 2: Short Papers)}, pages 37--42, Taipei, Taiwan. Asian Federation of Natural Language Processing.

\bibitem[{Pan et~al.(2020)Pan, Chen, Wu, Liu, Ngo, Kan, Jiang, and Chua}]{pan-etal-multi-2020}
Liang-Ming Pan, Jingjing Chen, Jianlong Wu, Shaoteng Liu, Chong-Wah Ngo, Min-Yen Kan, Yugang Jiang, and Tat-Seng Chua. 2020.
\newblock \href {https://doi.org/10.1145/3394171.3413765} {Multi-modal cooking workflow construction for food recipes}.
\newblock In \emph{Proceedings of the 28th ACM International Conference on Multimedia}, MM '20, page 1132–1141, New York, NY, USA. Association for Computing Machinery.

\bibitem[{Pareti et~al.(2014)Pareti, Testu, Ichise, Klein, and Barker}]{pareti2014integrating}
Paolo Pareti, Benoit Testu, Ryutaro Ichise, Ewan Klein, and Adam Barker. 2014.
\newblock Integrating know-how into the linked data cloud.
\newblock In \emph{International Conference on Knowledge Engineering and Knowledge Management}, pages 385--396. Springer.

\bibitem[{Park et~al.(2020)Park, Kulikov, Bhagavatula, and Choi}]{park-visualcomet-2020}
Jae~Sung Park, Ilia Kulikov, Chandra Bhagavatula, and Yejin Choi. 2020.
\newblock \href {https://arxiv.org/abs/2003.12506} {Visualcomet: Reasoning about the dynamic context of a still image}.
\newblock In \emph{European Conference on Computer Vision ({ECCV})}.

\bibitem[{Shangguan et~al.(2024)Shangguan, Qiu, Barbu, Fouhey, and Huang}]{shangguan-tomato-2024}
Ziyao Shangguan, Jiarui Qiu, Andrei Barbu, David Fouhey, and Song{-}Chang Huang. 2024.
\newblock \href {http://arxiv.org/abs/2410.23266} {{TOMATO}: Assessing visual temporal reasoning capabilities in multimodal foundation models}.
\newblock \emph{arXiv preprint}, arXiv:2410.23266.

\bibitem[{Valmeekam et~al.(2023{\natexlab{a}})Valmeekam, Marquez, Olmo, Sreedharan, and Kambhampati}]{valmeekam-etal-2023-planbench}
Karthik Valmeekam, Matthew Marquez, Alberto Olmo, Sarath Sreedharan, and Subbarao Kambhampati. 2023{\natexlab{a}}.
\newblock \href {https://arxiv.org/abs/2206.10498} {Planbench: An extensible benchmark for evaluating large language models on planning and reasoning about change}.
\newblock \emph{Preprint}, arXiv:2206.10498.

\bibitem[{Valmeekam et~al.(2023{\natexlab{b}})Valmeekam, Talamadupula, and Srivastava}]{valmeekam-planbench-2023}
Karthik Valmeekam, Kartik Talamadupula, and Sanjay Srivastava. 2023{\natexlab{b}}.
\newblock \href {https://openreview.net/forum?id=wUU-7XTL5XO} {{PlanBench}: An extensible benchmark for evaluating large language models on planning and reasoning about change}.
\newblock In \emph{NeurIPS Datasets \& Benchmarks Track}.

\bibitem[{Wei et~al.(2022)Wei, Wang, Schuurmans, Bosma, brian ichter, Xia, Chi, Le, and Zhou}]{wei-etal-2022-chain}
Jason Wei, Xuezhi Wang, Dale Schuurmans, Maarten Bosma, brian ichter, Fei Xia, Ed~H. Chi, Quoc~V Le, and Denny Zhou. 2022.
\newblock \href {https://openreview.net/forum?id=_VjQlMeSB_J} {Chain of thought prompting elicits reasoning in large language models}.
\newblock In \emph{Advances in Neural Information Processing Systems}.

\bibitem[{Wu et~al.(2022)Wu, Spangher, Alipoormolabashi, Freedman, Weischedel, and Peng}]{wu-etal-2022-understanding}
Te-Lin Wu, Alex Spangher, Pegah Alipoormolabashi, Marjorie Freedman, Ralph Weischedel, and Nanyun Peng. 2022.
\newblock \href {https://doi.org/10.18653/v1/2022.acl-long.310} {Understanding multimodal procedural knowledge by sequencing multimodal instructional manuals}.
\newblock In \emph{Proceedings of the 60th Annual Meeting of the Association for Computational Linguistics (Volume 1: Long Papers)}, pages 4525--4542, Dublin, Ireland. Association for Computational Linguistics.

\bibitem[{Yagcioglu et~al.(2018)Yagcioglu, Erdem, Erdem, and Ikizler{-}Cinbis}]{yagcioglu-recipeqa-2018}
Semih Yagcioglu, Aykut Erdem, Erkut Erdem, and Nazli Ikizler{-}Cinbis. 2018.
\newblock \href {https://arxiv.org/abs/1810.06553} {{RecipeQA}: A challenge dataset for multimodal comprehension of cooking recipes}.
\newblock In \emph{Proceedings of the 2018 Conference on Empirical Methods in Natural Language Processing ({EMNLP})}, pages 1358--1368.

\bibitem[{Yeo et~al.(2018)Yeo, Lee, Wang, Choi, Cho, Kim~Amplayo, and Hwang}]{yeo-etal-2018-visual}
Jinyoung Yeo, Gyeongbok Lee, Gengyu Wang, Seungtaek Choi, Hyunsouk Cho, Reinald Kim~Amplayo, and Seung-won Hwang. 2018.
\newblock \href {https://aclanthology.org/L18-1316/} {Visual choice of plausible alternatives: An evaluation of image-based commonsense causal reasoning}.
\newblock In \emph{Proceedings of the Eleventh International Conference on Language Resources and Evaluation ({LREC} 2018)}, Miyazaki, Japan. European Language Resources Association (ELRA).

\bibitem[{Yue et~al.(2024)Yue, Ni, Zhang, Zheng, Liu, Zhang, Stevens, Jiang, Ren, Sun et~al.}]{yue2024mmmu}
Xiang Yue, Yuansheng Ni, Kai Zhang, Tianyu Zheng, Ruoqi Liu, Ge~Zhang, Samuel Stevens, Dongfu Jiang, Weiming Ren, Yuxuan Sun, et~al. 2024.
\newblock Mmmu: A massive multi-discipline multimodal understanding and reasoning benchmark for expert agi.
\newblock In \emph{Proceedings of the IEEE/CVF Conference on Computer Vision and Pattern Recognition}, pages 9556--9567.

\bibitem[{Zellers et~al.(2019)Zellers, Holtzman, Bisk, Farhadi, and Choi}]{zellers-etal-2019-hellaswag}
Rowan Zellers, Ari Holtzman, Yonatan Bisk, Ali Farhadi, and Yejin Choi. 2019.
\newblock \href {https://doi.org/10.18653/v1/P19-1472} {{H}ella{S}wag: Can a machine really finish your sentence?}
\newblock In \emph{Proceedings of the 57th Annual Meeting of the Association for Computational Linguistics}, pages 4791--4800, Florence, Italy. Association for Computational Linguistics.

\bibitem[{Zhang et~al.(2020)Zhang, Chen, Wang, Song, and Roth}]{zhang-etal-2020-analogous}
Hongming Zhang, Muhao Chen, Haoyu Wang, Yangqiu Song, and Dan Roth. 2020.
\newblock \href {https://doi.org/10.18653/v1/2020.emnlp-main.119} {Analogous process structure induction for sub-event sequence prediction}.
\newblock In \emph{Proceedings of the 2020 Conference on Empirical Methods in Natural Language Processing (EMNLP)}, pages 1541--1550, Online. Association for Computational Linguistics.

\bibitem[{Zhang et~al.(2023)Zhang, Xu, Yang, Zhou, You, Arora, and Callison-Burch}]{zhang-etal-2023-causal}
Li~Zhang, Hainiu Xu, Yue Yang, Shuyan Zhou, Weiqiu You, Manni Arora, and Chris Callison-Burch. 2023.
\newblock \href {https://doi.org/10.18653/v1/2023.findings-eacl.31} {Causal reasoning of entities and events in procedural texts}.
\newblock In \emph{Findings of the Association for Computational Linguistics: EACL 2023}, pages 415--431, Dubrovnik, Croatia. Association for Computational Linguistics.

\bibitem[{Zhang et~al.(2025)Zhang, Hu, Lee, Shi, Kordjamshidi, Chai, and Ma}]{zhang2025do}
Zheyuan Zhang, Fengyuan Hu, Jayjun Lee, Freda Shi, Parisa Kordjamshidi, Joyce Chai, and Ziqiao Ma. 2025.
\newblock \href {https://openreview.net/forum?id=84pDoCD4lH} {Do vision-language models represent space and how? evaluating spatial frame of reference under ambiguities}.
\newblock In \emph{The Thirteenth International Conference on Learning Representations}.

\bibitem[{Zheng et~al.(2024)Zheng, Kant, Wang, Liu, Michael, and Iyyer}]{zheng-naturalplan-2024}
Huaixiu~Steven Zheng, Neel Kant, Rui Wang, Peter~J. Liu, Julian Michael, and Mohit Iyyer. 2024.
\newblock \href {http://arxiv.org/abs/2406.04520} {{NaturalPlan}: Benchmarking {LLMs} on natural-language planning}.
\newblock \emph{arXiv preprint}, arXiv:2406.04520.

\bibitem[{Zhou et~al.(2018)Zhou, Xu, Corso, Socher, and Xiong}]{zhou-youcook2-2018}
Luowei Zhou, Chenliang Xu, Jason~J. Corso, Richard Socher, and Caiming Xiong. 2018.
\newblock \href {https://openaccess.thecvf.com/content_cvpr_2018/papers/Zhou_Towards_Automatic_Learning_CVPR_2018_paper.pdf} {Towards automatic learning of procedures from web instructional videos}.
\newblock In \emph{Proceedings of the {IEEE}/CVF Conference on Computer Vision and Pattern Recognition (CVPR)}, pages 5832--5840.
\newblock Introduces the YouCook2 dataset.

\bibitem[{Zhukov et~al.(2019)Zhukov, Alayrac, Cinbis, Fouhey, Laptev, and Sivic}]{Zhukov2019}
Dimitri Zhukov, Jean-Baptiste Alayrac, Ramazan~Gokberk Cinbis, David Fouhey, Ivan Laptev, and Josef Sivic. 2019.
\newblock Cross-task weakly supervised learning from instructional videos.
\newblock In \emph{CVPR}.

\end{thebibliography}

\clearpage

\appendix

\section{Detailed Related Work}
\label{appendix:related-work}

Plan understanding tasks span testing knowledge and reasoning about multiple aspects such as entity states \cite{Bosselut2018SimulatingAD, henaff2017tracking},  actions \cite{pareti2014integrating, lin-etal-2020-recipe, donatelli-etal-2021-aligning}, next events \cite{nguyen-etal-2017-sequence,zellers-etal-2019-hellaswag,zhang-etal-2020-analogous} and more.
XPAD \cite{dalvi-etal-2019-everything} extend ProPara \cite{dalvi-etal-2018-tracking}, originally focused on understanding scientific processes, by adding the new task of explaining actions by predicting their dependencies. 
PizzaCommonsense \cite{diallo-etal-2024-pizzacommonsense} contains commonsense knowledge about intermediate and implicit steps for cooking recipes, providing explicit input/output pairs of each action with fine-grained annotations.
PlanBench \cite{valmeekam-planbench-2023} focuses on classical AI planning domains, such as BlocksWorld, and shows that \llms fail to produce valid, executable action sequences.
NaturalPlan \cite{zheng-naturalplan-2024} evaluate real-world constraint optimization based planning tasks.
CREPE \cite{zhang-etal-2023-causal} measures how well \llms understand the comparative likelihood of two events occurring in a procedure.
\catbench \cite{lal-catbench-2024} and \citet{kiddon-etal-2015-mise} explore predicting dependencies in cooking recipes.
Most work has evaluated how well models understand aspects of plans described only in a single (text) modality.

VCOPA \cite{yeo-etal-2018-visual} posit the task of identifying the correct next step given a premise step, with all the steps described in images.
VisualComet \cite{park-visualcomet-2020} present actions in stories as images along with summaries of what happened before and after, and show that integration between visual and textual commonsense reasoning is required to holistically reason about actions.
RecipeQA \cite{yagcioglu-recipeqa-2018} align text instructions in cooking recipes with images corresponding to the central action in each step.
TOMATO \cite{shangguan-tomato-2024} evaluates how multimodal foundational models perform temporal reasoning about continuous actions in videos.
\citet{pan-etal-multi-2020} build MM-ReS, the first large-scale dataset for cooking workflow construction, consisting of 9,850 recipes with human-labeled workflow graphs.
Similarly, the COMKitchens \cite{comkitchens_eccv2024} dataset provides cooking videos annotated with a structured visual action graph.
\citet{wu-etal-2022-understanding} test whether models can correctly sequence images of misordered actions described in instruction manuals.
COMFORT \cite{zhang2025do} assess \vlms' spatial reasoning capabilities of along the lines of frames of references.

Existing datasets evaluate different aspects of plans, but only a few focus on assessing whether models understand temporal ordering constraints on the steps of an instructional plan.
However, all these benchmarks only consider plans written in text. 
\iso is created to test using and integrating visual and textual understanding capabilities required to understand plans.

\section{Processing YouCook2}
\label{appendix:youcook2}

YouCook2 \cite{zhou-youcook2-2018} contains 2,000 cooking videos with clean stepwise captions and gives us YouTube links plus the exact time spans and step-by-step instructions for each recipe. 
We start by taking every span, jumping to its midpoint, and saving one JPEG frame; that leaves us with a clean image for every recipe step.

After we extracted a JPEG for every recipe step, we built a labeled pool of text-image pairs. We filter for videos with atleast 5 steps. For each such video it randomly picked an integer k between 2 and 4. If the task was ``before'', the text snippet is formed from the last k steps of the recipe; if the task was ``after'', the snippet uses the first k steps. We define two candidate frame groups:
(1) Positive pool – frames whose step comes just outside the snippet, within three steps:
For the ``before'' task these are the steps that immediately precede the snippet (i.e., recipe positions total-k-3 ... total-k-1).
For the “after” task they are the steps that immediately follow the snippet (k ... k+2).
These frames genuinely belong before or after the snippet, so pairing them with the snippet should be logically correct.
(2) Negative pool – frames taken inside the snippet itself (for ``before'', positions inside the last k steps; for ``after'', positions inside the first k steps). Because they come from the same span that forms the snippet, showing them separately violates the required temporal relation and therefore constitutes an incorrect match.

\begin{figure*}
    \centering
    \includegraphics[width=\textwidth]{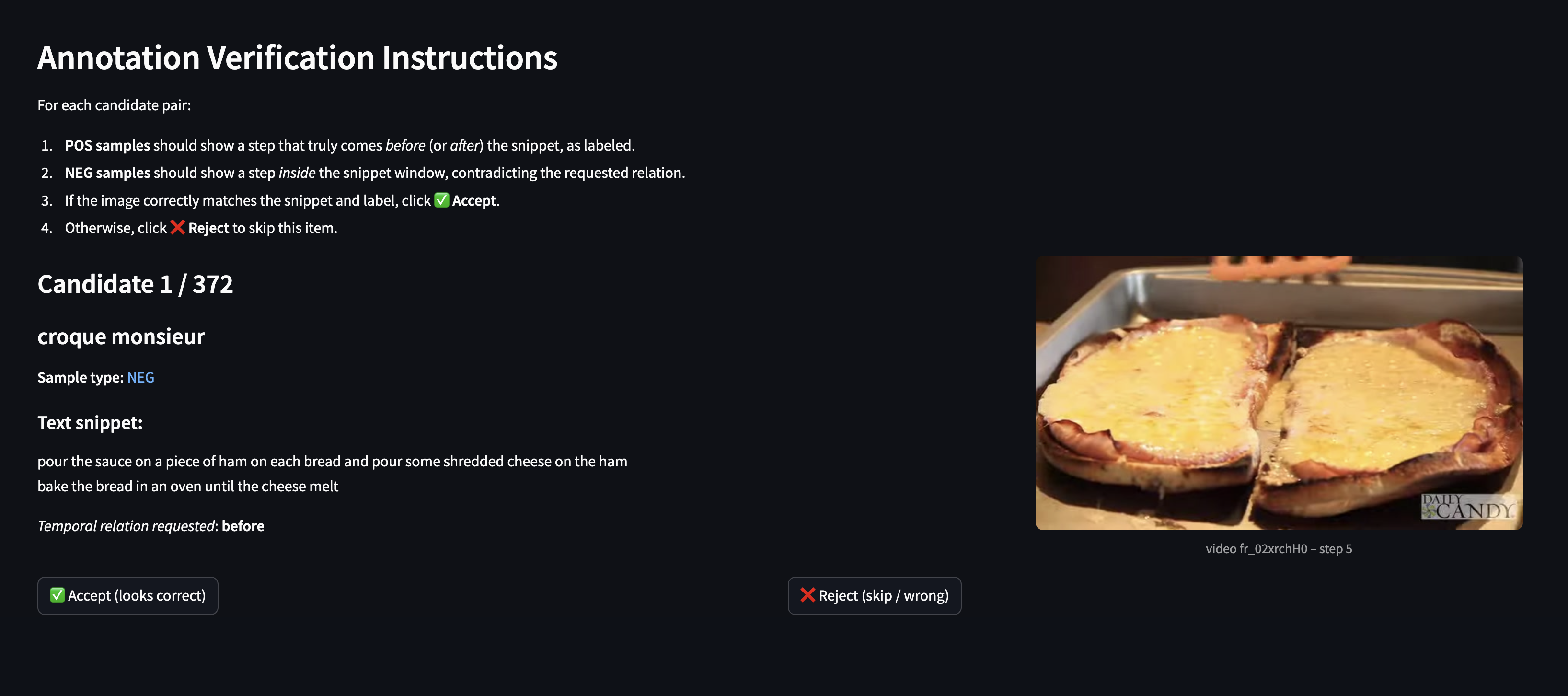}
    \caption{Screenshot with instructions provided to annotators from whom we collected gold labels for \iso.}
    \label{fig:annotation}
\end{figure*}

We built a Streamlit app to annotate each text-image pair with a gold label. The app shows the snippet, the frame, and whether the sample is tagged POS (the frame should come just before or after the snippet, as specified) or NEG (the frame is taken from inside the snippet and therefore should not match the requested relation). We look at the pair: if the frame truly matches (for a POS) or truly conflicts (for a NEG), we click Accept; otherwise we click Reject. 
We provide a screenshot of the annotation app in \autoref{fig:annotation}.

\section{Processing CrossTask}
\label{appendix:crosstask}

Starting from the 18 primary tasks, we capped ourselves at ten annotated clips per task (180 videos total).  For every step span in each clip we jumped to the midpoint and saved one JPEG, giving us a clean image for every unique step.

We perform the same procedure on data from CrossTask as we did in \autoref{appendix:youcook2}.

Using 180 videos we ended up with 139 human-vetted instances: 72 ``before'' and 67 ``after'', spanning 73 dependent/positive and 66 non-dependent/negative examples.

\section{Dataset Statistics}
\label{appsec:data_stats}

\begin{table}[h!]
\centering
\begin{tabular}{c r r r}
\toprule
Relation & \dep & \ndep & \iso \\
\midrule
Before            & 187    & 185       & 372  \\
After             & 196    & 196       & 392  \\
\midrule
\textbf{Total} & \textbf{383} & \textbf{381} & \textbf{764} \\
\bottomrule
\end{tabular}
\caption{\iso statistics with data points spanning 480 different instructional plan videos. Each data point contains a text snippet of the plan, an image description of a step not in the snippet, a binary question and answer about the causal dependence between them.}
\label{tab:data}
\end{table}

\section{Human Baseline for \iso}
\label{appsec:human_baseline}

To establish how well humans perform on this task, we presented the goal of the plan, the plan snippet and the image of a step and asked one annotator to answer the associated question.
We randomly sampled 200 data points from \iso for this, and we use the same task formulation that we use with models.
The last row in \autoref{tab:main_results} presents the per-class as well as macro average precision, recall and F1 scores for humans.
This establishes a possible upper bound for the \iso task.
We find that humans are able to identify implicit causal dependencies between steps represented across modalities fairly easily.
In fact, they rarely make mistakes.

\section{Benchmarking Models}
\label{appsec:benchmark_models}

We provide details of each model we evaluate on \iso.

\paragraph{\texttt{gpt-4o-2024-11-20}} accepts as input any combination of text, audio, image, and video and generates any combination of text, audio, and image outputs.
It is especially better at vision and audio understanding compared to existing models.

\paragraph{\texttt{gpt-4o-mini-2024-07-18}} has a context window of 128K tokens, supports up to 16K output tokens per request.
It surpasses other small models released to that date on academic benchmarks across both textual intelligence and multimodal reasoning, and supports the same range of languages as \gfomni.

\paragraph{\texttt{claude-3-5-sonnet-20241022}} sets new industry benchmarks for graduate-level reasoning (GPQA), undergraduate-level knowledge (MMLU), and coding proficiency (HumanEval). 
It shows marked improvement in grasping nuance, humor, and complex instructions, and is exceptional at writing high-quality content with a natural, relatable tone. 

\paragraph{\texttt{o3-2025-04-16}} excels at solving complex math, coding, and scientific challenges while demonstrating strong visual perception and analysis. It uses tools in its chains of thought to augment its capabilities; for example, cropping or transforming images, searching the web, or using Python to analyze data during the thought process.

\paragraph{\texttt{o4-mini-2025-04-16}} is a smaller model optimized for fast, cost-efficient reasoning—it achieves remarkable performance for its size and cost, particularly in math, coding, and visual tasks. It is the best-performing benchmarked model on AIME 2024 and 2025. It performs especially strongly at visual tasks like analyzing images, charts, and graphics.

\paragraph{\texttt{llava-1.5-7b-hf}} \cite{liu-llava-2023} is an open-source chatbot trained by fine-tuning LLaMA/Vicuna on GPT-generated multimodal instruction-following data. 
It is an auto-regressive language model, based on the transformer architecture.
It combines a frozen CLIP/Vicuna vision encoder with a 7B-parameter language backbone, and is instruction-tuned on about 600k image--text pairs and is open-source.

\paragraph{\texttt{Llama-3.2-11B-Vision-Instruct}} is the 11B version of the Llama 3.2-Vision set of multimodal LLMs which have been instruction tuned for image reasoning. It is built on top of the pretrained Llama 3.1 text only LLM by combining a seperately trained vision adapter module. Using a combination of supervised fine-tuning and reinforcement learning from human feedback, the model has been optimized to do a variety of vision tasks like image recognition, reasoning, captioning, and question answering on images.

\paragraph{\texttt{Qwen2.5-VL-32B-Instruct}} is a 32 billion parameter vision language model. It is created on top of the Qwen-2.5 7 billion language model by following the ViT architecture. It has been extensively instruction tuned on (image,text) tuples to so that the model understands all things visual, is agentic, can comprehend long videos and events, can do visual localization, and generate structured outputs.

\paragraph{\texttt{paligemma2-10b-mix-448}} checkpoints are fine-tuned on a diverse set of tasks and are ready to use out of the box while pt checkpoints are pre-trained and intended for further fine-tuning. These tasks include short and long captioning, optical character recognition, question answering, object detection and segmentation, and more. The model is available in the bfloat16 format for research purposes only.

\paragraph{\texttt{deepseek-vl2-small}} is a 16 billion parameters mixture-of-experts vision language model. It has shown been to demonstrate enhanced performance across multiple tasks like visual question answering, optical character recognition, document/table/chart understanding, and visual grounding. It improves upon its predecessor, DeepSeek-VL, by using an improved high-resoultion vision encoder for better visual comprehension and an optimized language model backbone for training and test time efficiency. It is trained on a data that boosts performance and gives new capabilities to the model such as precise visual grounding.

\section{Model Setup}
\label{appsec:model_setup}
Python was the main scripting language for data collection and experimentation.
For experiments using closed source models, we used OpenAI\footnote{\url{https://openai.com/api/pricing/}} and Anthropic\footnote{\url{https://://www.anthropic.com/pricing}} APIs. The total cost for OpenAI was $\sim$ 75 USD and  $\sim$ 20 USD for Claude. 
The open-source experiments were conducted on $4$ A6000 GPUs, each having $48$ GB. The total GPU hours for all the experiments was $\sim$15. 
The models were downloaded from Huggingface and hosted for inference using Huggingface transformers module and vLLM. 
We use greedy decoding and temperature set to $0.0$ when the option is available.
We use GitHub Co-Pilot to help with writing code but verify it manually before running any experiments.

\section{Prompts Used}
\label{appsec:prompts}

For each ISO instance, we provide the following prompt:

\begin{flushleft}\small\ttfamily
GOAL: \{\textit{Goal Name}\}\\
Plan:\\
\{\textit{step\_1}\}\\
\{\textit{step\_2}\}\\
\ldots\\
QUESTION: \{\textit{Does this image show a step that must come <before|after> the <first|last> step in the plan?}\}
\end{flushleft}

The prompt mirrors the reasoning process we want the model to follow.
The \texttt{GOAL} line primes topical knowledge about the dish (e.g.,
Indian curries usually add tomatoes after onions).  
The \texttt{Recipe excerpt} gives an ordered context without revealing
the full plan, so the model must place the image relative to these
steps—exactly the causal inference ISO is designed to test.
Finally, an explicit \texttt{QUESTION} asks about \emph{before} or
\emph{after}, limiting the required output to a binary choice and
removing ambiguity about expected format.

We test two modes:

\begin{enumerate}
    \item \textbf{Answer-Only}: We append ``Answer only with YES or NO.'' and use the first ``YES'' or ``NO'' as the prediction.
    \item \textbf{Explain-Then-Answer}: We request a short chain-of-thought enclosed by \texttt{<think>...</think>}, then a final answer enclosed by \texttt{<answer>YES</answer>} or \texttt{<answer>NO</answer>}. We evaluate only the final tag, similar to CoT prompting.
\end{enumerate}

The image is provided via the model's vision interface alongside this prompt. We limit generation to 120 tokens, which suffices for both reasoning and answer.

\paragraph{Metric.}
Since each instance is a binary question, we report \textit{accuracy}, \textit{precision}, \textit{recall}, and \textit{F1} on the subset of responses containing an unambiguous ``YES'' or ``NO.''\footnote{We discard outputs lacking either token.}

\end{document}